# Domestic waste detection and grasping points for robotic picking up

1st Victor De Gea
AUROVA Lab
Computer Science Research Institute
University of Alicante
San Vicente del Raspeig, Spain
victor.degea@ua.es

2nd Santiago T. Puente
AUROVA Lab
Computer Science Research Institute
University of Alicante
San Vicente del Raspeig, Spain
santiago.puente@ua.es

3rd Pablo Gil
AUROVA Lab
Computer Science Research Institute
University of Alicante
San Vicente del Raspeig, Spain
pablo.gil@ua.es

*Abstract*—This paper presents an AI system applied to location and robotic grasping. Experimental setup is based on a parameter study to train a deep-learning network based on Mask-RCNN to perform waste location in indoor and outdoor environment, using five different classes and generating a new waste dataset. Initially the AI system obtain the RGBD data of the environment, followed by the detection of objects using the neural network. Later, the 3D object shape is computed using the network result and the depth channel. Finally, the shape is used to compute grasping for a robot arm with a two-finger gripper. The objective is to classify the waste in groups to improve a recycling strategy.

## I. EXTENDED ABSTRACT

The objective of this work is to carry out a robotic manipulation of waste in order to classify it. To accomplish this task, a method which uses a RealSense d435i camera to obtain RGBD data from indoor/outdoor scenarios with waste is proposed. The RGB images have a spatial resolution of 640x480 pixels. They are used to feed the Mask-RCNN [1] to perform instances segmentation of objects in the scene.

It is crucial that the training dataset has sufficient representative samples for each type of object in order to obtain generic and discriminant features and to avoid the overfitting. For this reason, the images of the dataset were taken from different perspectives, objects appear with different positions, orientations, distances and visibility. Thus, two datasets with a total of 1434 images were created: one contains samples of waste in a indoor environment, and the other contains samples in an outdoor environment. Every image of these dataset is composed by a single object instance. A set of 27 waste objects were used in datasets generation. In these datasets, every image has been manually labeled, according to the following five classes to perform its classification: opaque plastic bottle, paperboard box, clear plastic bottle, drink can and opaque plastic container. The quantity of objects are balanced in the dataset

The mentioned neural network model allows to knew which pixels belong to the detected object and which do not. In this way, it is possible to perform an accurate detection of the objects on 2D images. The present approach consists of loading the pre-trained weights and then do a fine-tuning using the created dataset to adjust the model to the target problem. The fine-tuning is done by means of a hyper-parametric adjustment. To do the evaluation of the hyper-parameters a study of their impact in the neural network has been performed. The objective is to find a good configuration of parameters that allows testing the network model.

Furthermore, an algorithm that takes into consideration a two-finger robotic gripper and the predictions of Mask-RCNN model has been done. This algorithm allows the objects to be manipulated by contact forces on two points on their surfaces.

The proposal method can be divided into four steps (Figure 1). Firstly, the RGBD information of scene is captured, providing RGB images and depth maps simultaneously. Secondly, predictions on RGB images are computed and a binary mask is obtained for each detected object. Thirdly, using RGB data, D channel, and binary mask images of scene the 3D reconstruction of each detected object is performed. Finally, once 3D representation point cloud of objects are obtained, Geograsp algorithm is called, to compute grasping data on each one [2].

A similar work will be presented in [3]. It performs a garbage classification. To classify objects they have used GarbageNet which only identify the object in the image. Nevertheless, our proposal uses Mask-RCNN. Another difference is that they use GPD grasp [4] to compute gripper pose instead of grasping point.

In order to evaluate the performance of the trained model, a validation and test datasets have been used from created datasets. The performance in terms of MS COCO Average Precision(AP) [5] was tested with different values for Jaccard index (IoU) obtaining and AP of 0,370 and 0.703, AP50 (AP for IoU threshold of 0.5) of 0.556 and 0.896 and AP75 (AP for IoU threshold of 0.75) of 0.417 and 0.851, for test and validation data respectively. Furthermore, the performance has been tested for various scales of object too, obtaining a values of APs (AP for small objects) of 0.059 and 0.289 for test and validation subset, as well as 0.374 and 0.720 for APm (AP for medium objects) and 0.511 and 0.766 for APl (AP for large

This research was funded by Spanish Government through the project RTI2018-094279-B-I00. Besides, computer facilities were provided by Valencian Government and FEDER through the IDIFEFER/2020/003.

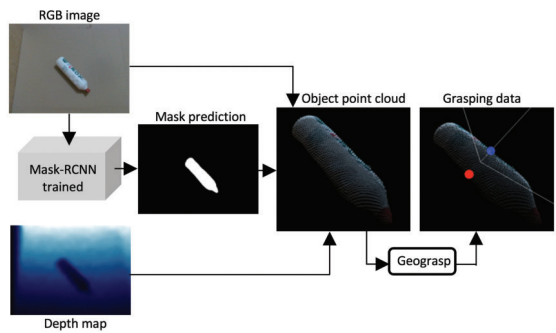

Fig. 1. Algorithm proposed.

objects). Taking this into consideration, the trained model has a higher precision on images similar to those used to train model (validation subset), while evaluating model with less similar images (test dataset), system returns a less level of precision.

Also, precision of trained model is evaluated on each category of object on test data obtaining in that trained model has a higher AP on opaque plastic bottles and opaque plastic containers. However, objects that return a less AP are clear plastic bottles. To observe the behaviour of the trained model from other point of view, some examples of predictions that model computes on images of test dataset are visualized (Figure 2). In some tests the predictions are incorrect, because the shadows and the reflection of the objects are classify like object pixels. While in other tests, class and mask predictions are executed correctly.

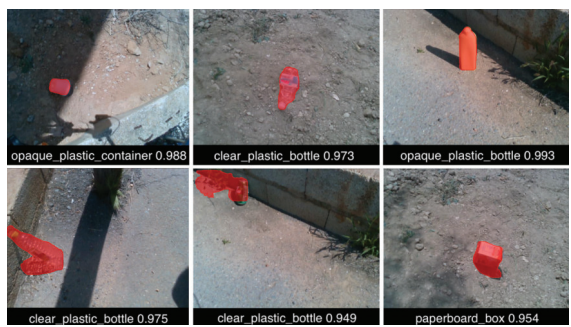

Fig. 2. Correct (first row) and incorrect (second row) predictions.

At this point, the grasping processing from mask predictions of trained model is evaluated. The experiments are carried out from correct mask detection of the objects, with the aim of evaluating only the processing of the grasping data.

Figure 3 shows grasping data compute for objects detected. Analyzing results, the fact of using a single camera on the processing allows to reconstruct only the area of the object visible by depth sensor. It can be seen that in (c) test the point cloud do not have enough 3D information about the object to calculate a good grip. This limitation depends on the positioning of the objects relative to the camera and shape of the objects. For clear objects depth data recorded are not quite correct, due to transparency. This causes that the reconstructions of these type of objects are not as accurate as in opaque objects. Good contact points are obtained on the tested object (f), but this limitations could affect the calculation of the grips. On the other hand, test (e) shows depth data of some pixels belonging to the drink can are not correct, which also could affect the grip calculation. Tests (a, b, d) shows correct grasping computing.

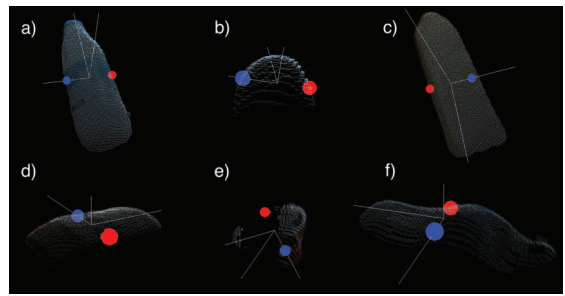

Fig. 3. Grasping data from mask predictions.

In conclusion, the study allows us to know a possible good parametric configuration based on created data, with which to train the model. In general, according to the tests performed, the proposed processing to calculate grasping data from predictions of the trained Mask-RCNN model allows obtaining correct grips. One point to highlight is that the system can be extend to a great number of classes in order to sort the waste using recycling criteria. Now the work is seting in a mobile platform in order to test in field the algorithm proposed. Furthermore, this extension to a field environment will allow us to improve the dataset and study the influence of a group of objects to grasp and the best sorting method to pick up all of them .